\documentclass[runningheads]{llncs}
\usepackage{graphicx}

\usepackage{tikz}
\usepackage{comment}
\usepackage{amsmath,amssymb} 
\usepackage{color}

\usepackage{graphicx}
\usepackage{amsmath}
\usepackage{amssymb}
\usepackage{booktabs}
\usepackage{multirow}
\usepackage[pagebackref,breaklinks,colorlinks]{hyperref}

\begin{document}
\pagestyle{headings}
\mainmatter
\def\ECCVSubNumber{4978}  

\title{Completing Partial Point Clouds with Outliers by Collaborative Completion and Segmentation} 


\titlerunning{CS-Net}
%
\author{Changfeng Ma\inst{1} \and
Yang Yang\inst{1} \and
Jie Guo\inst{1} \and
Chongjun Wang\inst{1} \and 
Yanwen Guo\inst{1}}
\authorrunning{C. Ma et al.}
%
\institute{National Key Lab for Novel Software Technology, Nanjing University}
\maketitle

\begin{abstract}
 Most existing point cloud completion methods are only applicable to partial point clouds without any noises and outliers, which does not always hold in practice. We propose in this paper an end-to-end network, named CS-Net, to complete the point clouds contaminated by noises or containing outliers. In our CS-Net, the completion and segmentation modules work collaboratively to promote each other, benefited from our specifically designed cascaded structure. With the help of segmentation, more clean point cloud is fed into the completion module. We design a novel completion decoder which harnesses the labels obtained by segmentation together with FPS to purify the point cloud and leverages KNN-grouping for better generation. The completion and segmentation modules work alternately share the useful information from each other to gradually improve the quality of prediction. To train our network, we build a dataset to simulate the real case where incomplete point clouds contain outliers. Our comprehensive experiments and comparisons against state-of-the-art completion methods demonstrate our superiority. We also compare with the scheme of segmentation followed by completion and their end-to-end fusion, which also proves our efficacy.
\end{abstract}

\section{Introduction}






Point clouds are the raw standard outputs of most 3D scanning devices~\cite{kim2018scan,chen2019unpaired}. In recent years, they have gained more popularity~\cite{guo2020deep} as a fundamental data structure to represent 3D data and to process 3D geometry~\cite{dai2017scannet,chen2017multi,li20173d}. However, incomplete point clouds could be frequently encountered in practice due to the nature of scanners and object occlusions during scanning, heavily hampering the downstream applications. Point cloud completion which infers a complete object model given a partial, incomplete point cloud thus is of important use and has received considerable attention. Many deep learning based approaches~\cite{yuan2018pcn,yang2018foldingnet,wang2020cascaded} have been developed so far including the TopNet~\cite{tchapmi2019topnet}, SnowflakeNet~\cite{xiang2021snowflakenet}, and PMP-Net~\cite{wen2021pmp}, {\em etc.} These methods show remarkable performance for recovering the original shapes from incomplete point clouds.

\begin{figure}
	\begin{center}
		\includegraphics[width=1\linewidth]{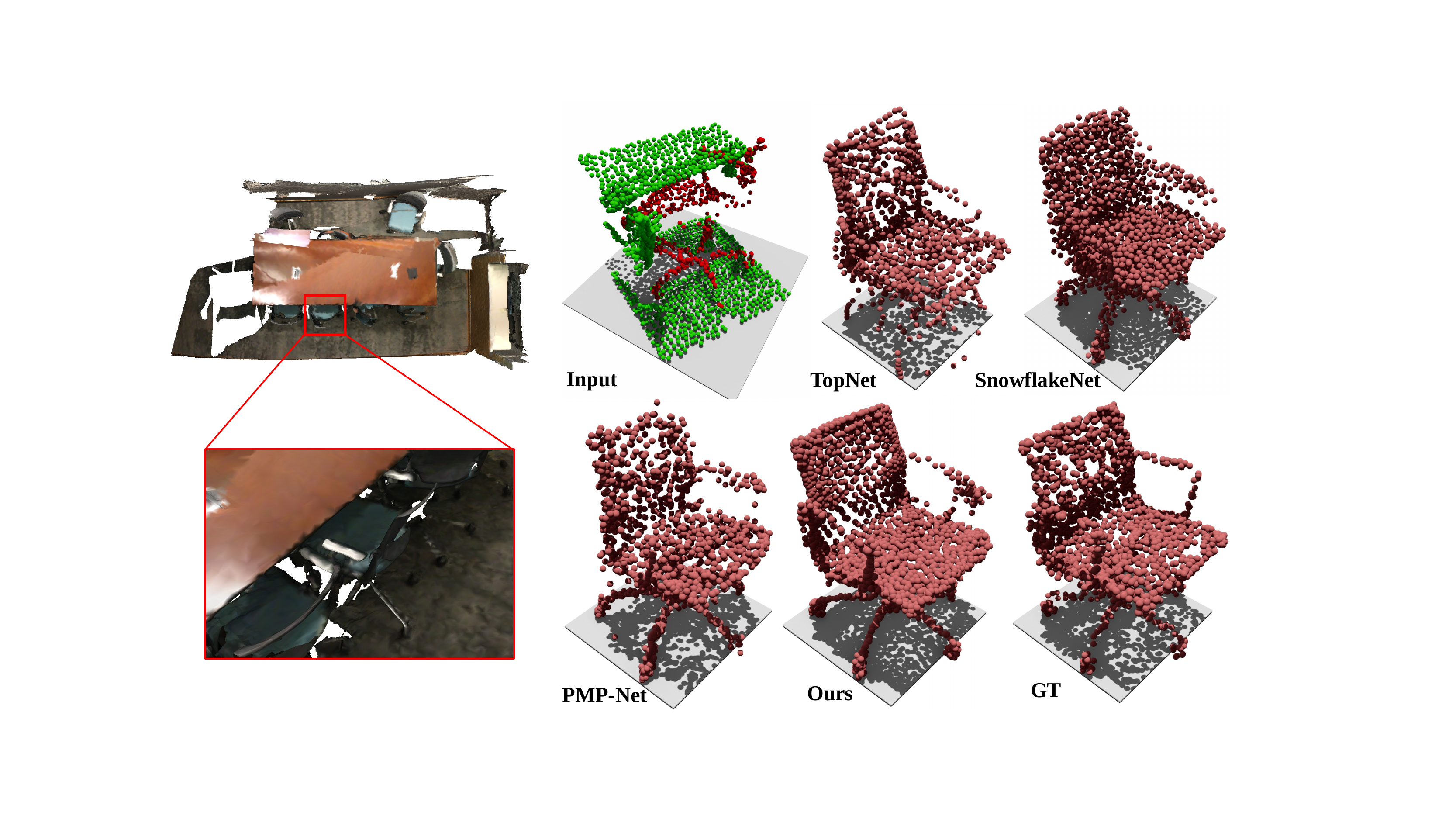} 
		\vspace{-4mm}
	\end{center}
	\caption{Comparison on point cloud completion results by different methods, including TopNet~\cite{tchapmi2019topnet}, SnowflakeNet~\cite{xiang2021snowflakenet}, PMP-Net~\cite{wen2021pmp}, and our CS-Net. As seen, previous methods generally fall short in completing the point cloud containing outliers which could be frequently encountered during scanning. By contrast, our SC-Net can generate the complete shape with clean and fine details. Green denotes outliers.}

	\label{Fig1-comp}
	\vspace{-4mm}
\end{figure}

Existed completion datasets are generated by directly sampling the shapes from 3D model datasets, thus previous methods generally assume that the input point clouds are clean and do not contain any noises and outliers. We argue that such an assumption does not always hold in practice. When scanning an object in the scene, the surroundings around the target object will be inevitably seen by the scanning devices. One example is shown in Fig.\ref{Fig1-comp}. Though semantic or instance segmentation, as a pre-processing step, has achieved promising performance, misclassification is inevitable. The state-of-the-art instance segmentation methods \cite{Liang_2021_ICCV_SSTNet,Chen_2021_ICCV_HAIS} report around 68\% average precision on the public ScanNet dataset~\cite{dai2017scannet} and S3DIS dataset~\cite{armeni_cvpr16_S3DIS}. That is to say, accurately cutting out the target point cloud model from the whole point cloud still remains challenging and is an unresolved issue. Under such a circumstance, for point cloud completion a practical situation we have to face in practice is the partial point cloud model to be completed inevitably contains severe noises or outliers, which would severely degrade the performance of completion. Figure \ref{Fig1-comp} shows the completion results by different methods for the partial point cloud containing outliers. Previous methods generally fall short in completing the partial point clouds with noise or outliers. 


In this paper, we propose a method for completing the partial object cloud from the input with severe noises or outliers, termed CS-Net. CS-Net is an end-to-end cascaded neural network, which takes partial point clouds with noisy points as inputs and outputs noise-free complete results. Completion and segmentation modules of our CS-Net work collaboratively by making use of the shape information learnt from each other, enabling mutual promotion. With the help of segmentation, the completion module is able to utilize the clean input point cloud for completion. Meanwhile, the segmentation module is able to distinguish noisy points from target objects more accurately with the help of the complete shape inferred by completion. We specifically design a novel completion decoder which harnesses the label information obtained from segmentation together with the farthest point sampling (FPS) to purify the input for completion. Furthermore, KNN-grouping is incorporated into the encoder to search for ideal point positions with high probabilities for better point generation. 
The cascaded architecture of CS-Net refines the results of segmentation and completion block by block and also passes on the features between the two modules. Our network is able to generate the detailed, complete point cloud and accurate segmentation results simultaneously. 

The current datasets for point cloud completion mainly comprise clean point clouds and are insufficient to simulate real cases. To train our CS-Net, we build a dataset to simulate the real case that contains both partial point clouds and noisy points and outliers from surrounding objects. The dataset, generated from ScanNet v2~\cite{dai2017scannet}, ShapeNet~\cite{chang2015shapenet} and Scan2CAD~\cite{Avetisyan_2019_CVPR_scan2cad}, will be made publicly available. We also retrain and evaluate representative SOTA segmentation and completion networks.

An alternative to our solution that sounds reasonable is to first cut out the partial target model via semantic or instance segmentation, and then complete the clean point cloud in an independent step. We experiment with these two-stage schemes, and the comparison further validates the effectiveness of our end-to-end unified solution. Our advantage could be due to the fact that the shape information learnt by the segmentation and completion modules could benefit from each other. By contrast, with two independent steps the inaccurate segmentation still poses significant challenges to previous completion methods. We also combine segmentation and completion networks to end-to-end networks for further comparison. The results show that our specially designed label-multiplication-FPS module can fuse the information from segmentation and completion, by which these two modules achieve mutual promotion.


The major contributions of our work are as follows.
\begin{itemize}

\item We propose, for the first time, an end-to-end network for completing point clouds contaminated by noises or containing outliers. With our network architecture, the segmentation and completion modules, working collaboratively, promote each other's performance by leveraging the shape information learnt by different branches.
\item We design a novel completion decoder which harnesses the labels obtained from segmentation together with FPS to purify the point cloud and leverages KNN-grouping to search for proposed points for better point generation.
\item We evaluate the performance of CS-Net through comprehensive experiments. Specifically, we also show our superiority by comparing against the two-stage scheme of segmentation followed by completion and their end-to-end fusion. 
\item We contribute to the community a new point cloud completion dataset to simulate the input that contains both partial point clouds and noisy points or outliers from surrounding objects. And we apply sufficient evaluation of previous completion networks on our dataset.

\end{itemize}

The rest of this paper is organized as follows. Section 2 reviews briefly the works on point cloud completion and segmentation. Section 3 introduces our CS-Net framework for point cloud completion from noisy input in detail. Dataset generation, experiments and comparisons are shown in Section 4 and Section 5. Section 6 concludes the whole paper and highlights future work.

\section{Related work}

We review here briefly previous works on deep learning based point cloud completion and segmentation.

{\bf Deep learning on point clouds.} Early works~\cite{stutz2018learning,sharma2016vconv} represent shapes using voxel grids in order to directly use 3D convolution for shape analysis. The pioneer works of deep learning on point clouds are PointNet~\cite{2017PointNet} and its extension~\cite{2017PointNet++}, which directly consume point clouds and well respect permutation invariance of points in the input. Li {\em et al.}~\cite{li2018pointcnn} propose a PointCNN algorithm which is a generalization of traditional CNNs for feature learning on point clouds. Moreover, DGCNN \cite{dgcnn} devises EdgeConv to extract features from point clouds. Later research works build upon above basic methods to extracrt features for tasks relating to point clouds, such as classification \cite{joseph2019momen,zhao2019pointweb,le2018pointgrid}, semantic segmentation \cite{jiang2018pointsift,te2018rgcnn,landrieu2019point,komarichev2019cnn}, up-sampling~\cite{li2019pu,yu2018pu,2018Learning}, and 3D reconstruction~\cite{li2018efficient,lin2017learning}.

{\bf Point cloud completion.} In the early stage, existing successful works~\cite{thanh2016field,dai2017shape} utilize voxels as representations of 3D models on which 3D convolutional neural networks can be immediately applied. 
Yuan {\em et al.}~\cite{yuan2018pcn} first utilize an encoder-decoder architecture to generate coarse complete outputs from partial point clouds, followed by up-sampling and refining for more detailed outputs. This coarse-to-fine completion framework has also been employed by~\cite{xie2020grnet,Zhang_2020_ECCV_NSFA}. 
TopNet~\cite{tchapmi2019topnet} and~\cite{liu2019morphing} design different new decoders to generate more structured complete point clouds from global features extracted by encoder. 
Huang {\em et al.}~\cite{huang2020pf} focus on predicting the missing part of the incomplete point cloud by utilizing a multi-resolution encoder and a point pyramid decoder.
Wang {\em et al.}~\cite{wang2020cascaded} first predict a coarse point cloud and then combine the input point cloud with it. After farthest point sampling~\cite{2017PointNet++}, the combination is refined by a cascaded reconstruction module. 
RL-GAN-Net~\cite{Sarmad_2019_CVPR_RL_GAN_Net} and~\cite{Xie_2021_CVPR_Style_Based} apply generative adversarial networks to complete the partial point clouds.
Wen {\em et al.}~\cite{wen2020point} combine the folding module with self-attention in a hierarchical decoder that makes use of features extracted from the multi-level encoder. 
New encoders are proposed by~\cite{wang2020Softpool} and~\cite{zhang2021pointsetvoting} to learn better local features from neighbor points for point cloud completion and other tasks such as segmentation. 
PMP-Net~\cite{wen2021pmp} generates complete point clouds by moving input points to appropriate positions iteratively with minimum moving distance.
Additional information such as images or edges of models is leveraged by~\cite{Zhang_2021_CVPR_View_Guided} and~\cite{Gong_2021_ICCV_ME-PCN} to complement with the point cloud input.
SnowflakeNet~\cite{xiang2021snowflakenet} utilizes transformer and point-wise feature deconvolutional modules to refine the first-stage point could with multiple times.
And Yu {\em et al.}~\cite{yu2021pointr} also utilize a transformer-based architecture which takes unordered groups of points with position embeddings.
Huang {\em et al.}~\cite{huang2021rfnet} propose an efficient and fast-converging recurrent forward network that decreases the memory cost by sharing parameters.
HyperPocket~\cite{spurek2021hyperpocket} first completes the point clouds with an autoencoder-based architecture and then leverages a hypernetwork to make point clouds adapted to the scenes which fill the holes in the scenes.

Though recent works have some remarkable performances on point cloud completion, they generally assume that the input partial model is clean, without any noises and outliers, which is an ideal circumstance rarely encountered in practice. The performance would be degraded given the point clouds with noises or outliers as inputs. 

{\bf Point cloud segmentation} is crucial and fundamental for 3D scene understanding. 
For semantic segmentation, Jiang {\em et al.}~\cite{2018PointSIFT} propose a multi-stage ordered convolution module to stack and encode the information from eight spatial achieve orientation encoding and scale awareness.
Li {\em et al.}~\cite{2020Complete} design sparse voxel completion network to complete the 3D surfaces of a sparse point cloud for promoting the performance of the semantic segmentation network.
Recent voxel-point-based semantic segmentation work AF$^2$-S3Net~\cite{2021-2-S3Net} presents a multi-branch attentive feature fusion encoder and an adaptive feature selection decoder with feature map re-weighting. 
SGPN~\cite{wang2018SGPN} and GSPN~\cite{li2019GSPN} separately propose clustering-based and proposal-based methods for point cloud instance segmentation. Recent work HAIS~\cite{Chen_2021_ICCV_HAIS} makes full use of the relation between points and point sets and generated instance results progressively by a hierarchical aggregation architecture. Liang {\em et al.}~\cite{Liang_2021_ICCV_SSTNet} split nodes of a pre-trained, intermediate, semantic superpoint tree for proposals of instance objects. 
Point Transformer~\cite{zhao2021point} introduces transformer in the encoding process of point cloud to learn the representation, which can be applied to different point cloud tasks such as segmentation and classification.

Although existing methods have achieved some impressive results, it is undoubted that accurate segmentation of target models without errors still remains challenging and unresolved. 






\section{Method}
\begin{figure*}[t]
	\begin{center}
		\includegraphics[width=1\linewidth]{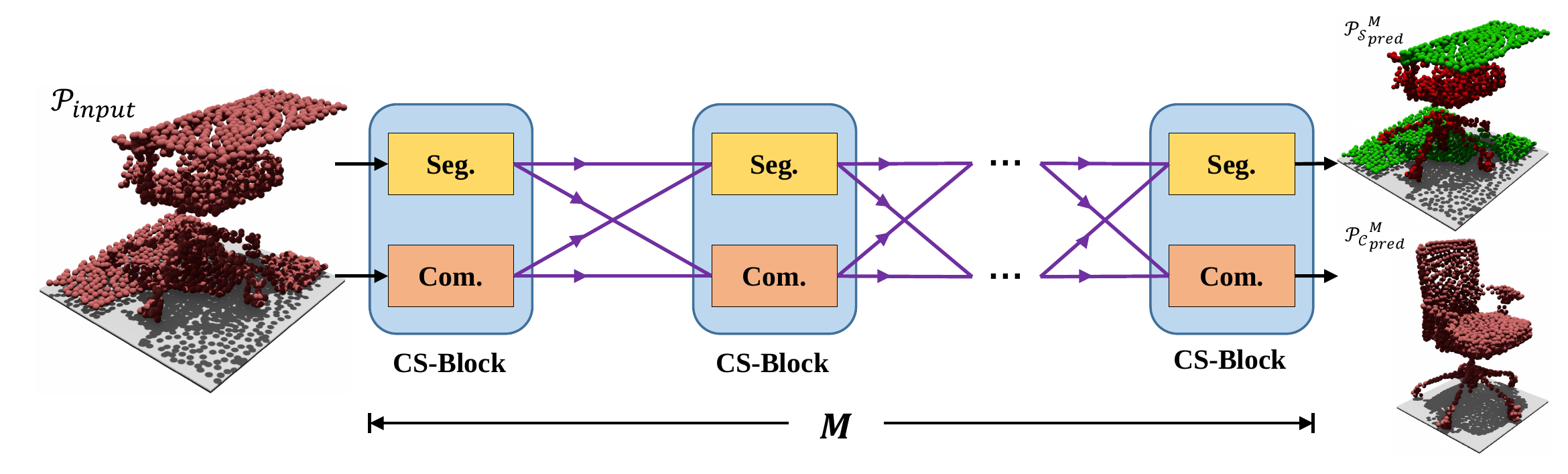}
	\end{center}
	\caption{The overall architecture of our CS-Net. The network has $M$ CS-Blocks, which consists of segmentation modules (Seg.) and completion modules (Com.). The segmentation module and completion module work collaboratively to refine the results in each block. The final outputs include the segmentation result and completed point cloud. Green indicates outliers.}
	\label{main_net}
\end{figure*}

We first explain the architecture of our CS-Net for the completion of partial point clouds with severe noises or outliers and then describe the loss functions used to train CS-Net. Figure \ref{main_net} shows the complete architecture of CS-Net.

\subsection{Network Architecture}







Our network is mainly comprised of several CS-Blocks which are responsible for simultaneous completion and segmentation. The input to the network is a point set $\mathcal{P}_{input} = \{(x_j, y_j, z_j)|j=1,2,...,N\}$ containing the partial point cloud with severe noises or outliers, where $N$ denotes the point number. Each CS-Block takes the outcomes and features from the last block and predicts the completion result $\mathcal{P_C}_{pred}^{i}$ and segmentation result $\mathcal{P_S}_{pred}^{i}\in [0, 1]^N$, where $i=1,2,...,M$.

\subsubsection{Overall architecture.}
The overall architecture makes segmentation and completion collaborate with each other. As shown in Figure~\ref{main_net}, the CS-Net contains $M$ CS-Blocks, each of which consists of a completion module and a segmentation module. The segmentation and completion modules pass their outcomes and features to both the segmentation and completion modules in the next block. As features passed from a CS-Block to the next CS-Block, features are shared between segmentation and completion modules, which achieves the collaboration of completion and segmentation. On the other hand, CS-Blocks can refine the outcomes from the last block. Since all the feature connections and transfer operations are differentiable, the segmentation module and completion module can influence and promote each other when training the network with back propagation.

\vspace{-2mm}
\subsubsection{Segmentation Module.} 
\begin{figure*}[t]
	\begin{center}
		\includegraphics[width=1\linewidth]{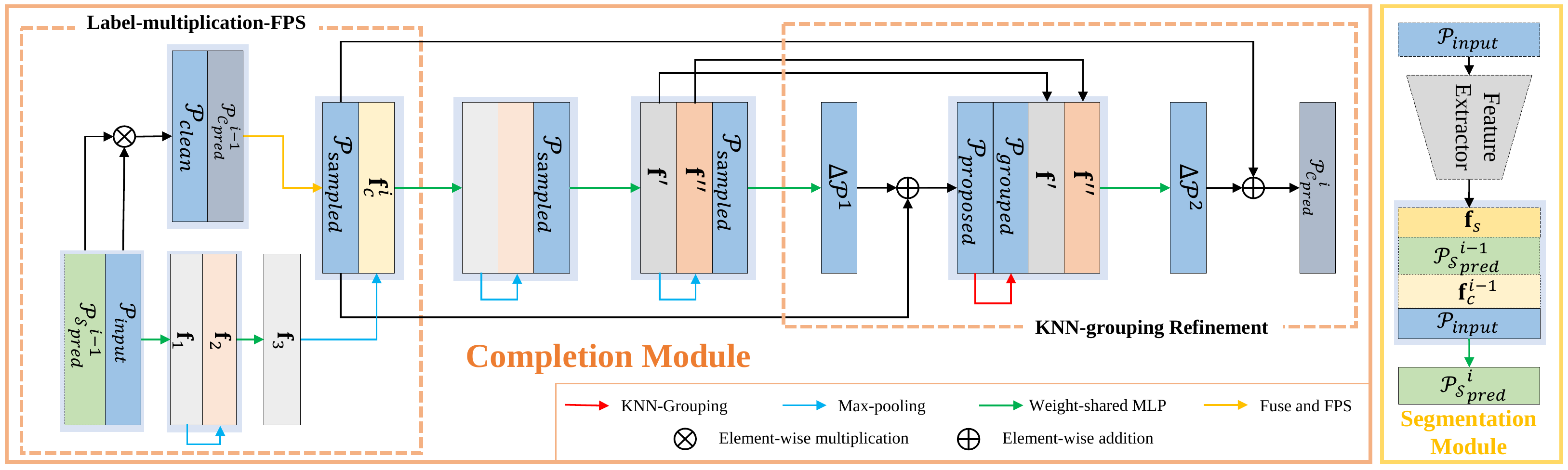}
	\end{center}
	\caption{Structure of the completion module and segmentation module. The completion module mainly consists of a label-multiplication-FPS module and a KNN-grouping refinement module. The label-multiplication-FPS module is responsible for removing noises and re-sampling the combination of clean input and output from the previous iteration. The KNN-grouping refinement module utilizes the intended position of points in a local area to generate new point clouds. The segmentation module mainly consists of a feature extractor and a MLP for predicting the labels.}
	\label{CD}
\end{figure*}

The segmentation module first extracts feature from input point cloud $\mathcal{P}_{input}$, then utilizes the feature and other additional features to predicts the labels $\mathcal{P_S}_{pred}^{i}\in [0, 1]^N$. For efficiency, the feature extractor only extract feature $\mathbf{f}_s$ once as show in the right of Figure \ref{CD}. $\mathbf{f}_s$ is then passed to next blocks and shared by all segmentation modules. The feature extractor here can be any network such as PointNet\cite{2017PointNet}, PointNet++~\cite{2017PointNet++}, Point Transfomer~\cite{zhao2021point} and so on. We apply a weight-shared MLP taking $\mathbf{f}_s$ from feature extractor, and $\mathcal{P}_{input}$ as input to directly predict the segmentation labels $\mathcal{P_S}_{pred}^{i}\in [0, 1]^N$. Such MLP in the $i$-th CS-Block, where $i>1$, also takes additional information including the $\mathbf{f}_{c}^{i-1}$ and the $\mathcal{P_S}_{pred}^{i-1}$ from the previous CS-Block as input.


\vspace{-3mm}
\subsubsection{Completion Module.} As shown in the left of Figure \ref{CD}, the $i$-th completion module, where $i>1$, utilizes the input point cloud and the outcomes $\mathcal{P_C}_{pred}^{i-1}$ and $\mathcal{P_S}_{pred}^{i-1}$ from the previous block for completing and refining the $\mathcal{P_C}_{pred}^{i-1}$ to get a better complete point cloud. And the first completion module utilize~\cite{tchapmi2019topnet} to generate the course point clouds.

{\bf Label-Multiplication-FPS Module.} To directly utilize the information from $\mathcal{P}_{input}$, we fuse $\mathcal{P_C}_{pred}^{i-1}$ with $\mathcal{P}_{input}$ to get details of objects from inputs. However, direct fusion may re-introduce noisy points, leading to completion errors. To solve this problem, we design a label-multiplication-FPS module to purify the fused point cloud. To remove noisy points in $\mathcal{P}_{input}$, we multiply it by the segmentation labels $\mathcal{P_S}_{pred}^{i-1}=\{l_j|j=1, 2, ...,N\}$. This in fact modulates the input with the fidelity yielded by segmentation and produces a purified point cloud $\mathcal{P}_{clean}$:
\begin{equation}
	\mathcal{P}_{clean} = \{(l_j\times x_j,l_j\times y_j,l_j\times z_j)|j=1,2,...,N\}.
\end{equation}
$\mathcal{P}_{clean}$ and $\mathcal{P_C}_{pred}^{i-1}$ are then fused together to get a point cloud with $2N$ points:
\begin{equation}
\mathcal{P}_{fused}=\mathcal{P}_{clean}\cup \mathcal{P_S}_{pred}^{i-1},
\end{equation}
where $\mathcal{P_C}_{pred}^{i-1}$ is the complete point cloud predicted by the previous CS-Block. By applying farthest point sampling (FPS)~\cite{2017PointNet++}, we remove the outliers around $(0, 0, 0)$ that are produced due to multiplication and obtain the sampled point cloud $\mathcal{P}_{sampled}$ with $N$ points.
The other part of label-multiplication-FPS module is responsible for extracting a global feature to guide the completion and be shared to segmentation modules. After extracting the features $\mathbf{f}_1$ from $\mathcal{P}_{input}$ and $\mathcal{P_S}_{pred}^{i-1}$ by a weight-shared MLP~\cite{2017PointNet}, we first utilize the max-pooling operation to obtain the global feature by maximizing each channel, and then repeat this feature to get $\mathbf{f}_2$. The next MLP takes the combination of $\mathbf{f}_1$ and $\mathbf{f}_2$ and outputs $\mathbf{f}_3$. We apply max-pooling to $\mathbf{f}_3$ to obtain $\mathbf{f}_{c}^i$, which offers global information to segmentation and completion modules at the same time. 

After passing through the label-multiplication-FPS module, the points in $\mathcal{P}_{sampled}$ are appended with $\mathbf{f}_{global}^i$ and passed through a module similar to the second part of label-multiplication-FPS module, where the input to each MLP also contains $\mathcal{P}_{sampled}$ as shown in the left of Figure \ref{CD}.

{\bf KNN-Grouping Refinement Module.} The positions of points gathering in a local area can be utilized to obtain a better position. To this end, we further design a KNN-grouping refinement module which takes the extracted features $\mathbf{f}'$ and $\mathbf{f}''$ with $\mathcal{P}_{sampled}$ for refinement. We first predict the shift $\Delta \mathcal{P}^1$ by employing a MLP taking the combination of $\mathbf{f}'$, $\mathbf{f}''$ and $\mathcal{P}_{sampled}$ and then add the $\Delta \mathcal{P}^1$ to the $\mathcal{P}_{sampled}$ to get the proposed point cloud $\mathcal{P}_{proposed}$. We apply a KNN search on $\mathcal{P}_{proposed}$ to get the indexes of neighbors according to which we group the points and get $\mathcal{P}_{grouped}$ with shape $N\times (K\times 3)$, where $K$ denotes the number of the neighbors belonging to each point. The input to the last MLP includes $\mathcal{P}_{proposed}$, $\mathcal{P}_{grouped}$, $\mathbf{f}'$ and $\mathbf{f}''$ and the output is a shift $\Delta \mathcal{P}^2$. Finally, the KNN-grouping refinement module outputs refined point cloud $\mathcal{P_C}_{pred}^i$ by adding $\Delta \mathcal{P}^2$ to $\mathcal{P}_{sampled}$.

\subsection{Loss Functions}
For effective and efficient learning, our network adopts the following loss functions.

{\bf Segmentation loss.} To measure the difference between the predicted segmentation labels and ground truth, we employ the cross-entropy loss:
\begin{equation}
	\begin{split}
		\mathcal {L}_{seg}(\mathcal{P_S}_{pred}, \mathcal{P_S}_{gt}) & =  -\frac{1}{| \mathcal{P_S}_{gt}|}\sum  \big[\mathcal{P_S}_{gt}\log(\mathcal{P_S}_{pred}) \\  & + (1-\mathcal{P_S}_{gt})\log(1-\mathcal{P_S}_{pred})\big],
	\end{split}
\end{equation}
where $\mathcal{P_S}_{pred}\in [0, 1]^N$ represents the predicted labels and $\mathcal{P_S}_{gt}\in \{0, 1\}^N$ denotes the ground truth labels.

{\bf Reconstruction loss.} Chamfer Distance (CD)~\cite{fan2017point} is a well-known loss function to measure the similarity between point clouds. It calculates the average closest points distance between the output point cloud $\mathcal{P}_{out}$ and the ground truth $\mathcal{P}_{gt}$. The loss value on a point $x$ compared with another point cloud $\mathcal{P}$ is the minimum distance between $x$ and all points in $\mathcal{P}$:
\begin{equation}
	\mathcal {L}_{CD}(x, \mathcal{P}) = \min_{y\in \mathcal{P}} ||x-y||_2.
\end{equation}
The CD loss between two point clouds $\mathcal{P}_1$ and $\mathcal{P}_2$ is:
\begin{equation}
	\begin{split}
		\mathcal {L}_{CD}(\mathcal{P}_1, \mathcal{P}_2) & = \frac{1}{|\mathcal{P}_1|}\sum_{x\in \mathcal{P}_1}\mathcal {L}_{CD}(x, \mathcal{P}_2)\\
		& + \frac{1}{|\mathcal{P}_2|}\sum_{y\in \mathcal{P}_2}\mathcal {L}_{CD}(y, \mathcal{P}_1).
	\end{split}
\end{equation}
The first term measures the distance between the output point cloud and ground truth, and the second one takes charge of the opposite direction.

{\bf Overall loss.} For $M$ output pairs $\{(\mathcal{P_S}_{pred}^i, \mathcal{P_C}_{pred}^i)|i = 1, 2, ..., M\}$ from our cascaded decoder and the ground truth $\mathcal{P_S}_{gt}$ and $\mathcal{P_C}_{gt}$, the overall loss is defined as,
\begin{equation}
    \mathcal{L} = \sum_{i=1}^M\left(\alpha_1 \mathcal{L}_{seg}(\mathcal{P_S}_{pred}^i, \mathcal{P_S}_{gt})+\alpha_2 \mathcal {L}_{CD}(\mathcal{P_C}_{pred}^i, \mathcal{P_C}_{gt})\right),
\end{equation}
where $\alpha_1$ and $\alpha_2$ are the weights used to balancing the influences of the segmentation loss and reconstruction loss. They are set to 0.01 and 1 separately during training.


\section{Dataset}\label{sec:dataset}

The current datasets on point cloud completion~\cite{yuan2018pcn,tchapmi2019topnet,huang2020pf} only consider the ideal case where the partial point clouds to be completed are free from noises and outliers. To simulate the real case where each sample contains both partial point clouds or outliers from surrounding objects, we use three datasets: ScanNet v2~\cite{dai2017scannet}, ShapeNet~\cite{chang2015shapenet} and Scan2CAD~\cite{Avetisyan_2019_CVPR_scan2cad} to synthesize a new dataset. We also retrain and evaluate existing representative methods on our dataset in three different ways.  

\subsection{Data Generation}
We first generate complete point clouds by sampling CAD models of ShapeNet, serving as the ground truths. Then we use annotations from Scan2CAD to align these complete point clouds with their corresponding objects in real scenes and remove the real-scanned objects at the same time. We then cut out the points in the boxes which are 10\% larger than the bounding boxes of aligned CAD models. Next, we use~\cite{Zhou2018open3d,sagi07vis-point,Mehra2010vis-noisy} to remove unseen points from the complete point clouds in two random views. Finally, we voxel-sampling and normalize the generated point clouds for effective learning. The labels of points sampled from CAD models are 1 and other points from surroundings are 0, which are the ground truth for segmentation. We choose 21,054 models of 10 categories, including chair, table, trash bin, TV or monitor, cabinet, bookshelf, sofa, lamp, bed and tub that are present in both the ScanNet and ShapeNet datasets. We split these models into the train, validation and test sets which contain 16,858,  2,096 and 2,100 models separately.

\subsection{Evaluation}
We retrain existing representative methods on our dataset in three different ways: using the original network, using a non-end-to-end framework, using an end-to-end framework. The first way is directly put the partial point clouds with noises into the original network. The second way is a two-stage scheme. We first train segmentation networks using the partial point clouds with noises of our dataset. We then train several previous methods using the partial point clouds where noises are removed according to the predicted label of the segmentation networks. In the first stage, we use segmentation model to label and remove outliers. Then in the second stage, we complete the resulting clean point clouds, using different state-of-the-art completion networks. The third way is to fuse the segmentation and completion networks together by feature sharing, leading to an end-to-end framework. We directly share the features of encoders and results between the segmentation and completion networks.

We use Chamfer Distance (CD)~\cite{fan2017point}, Density-aware Chamfer Distance (DCD)~\cite{Wu2021DensityawareCD}, F-Score@0.01\% (F$^{0.01\%}_{score}$) and F-Score@0.1\% (F$^{0.1\%}_{score}$)~\cite{tatarchenko2019single} to evaluate the complete point clouds. The evaluation results are shown in Section \ref{sec:exp}.

\begin{figure*}[t]
	\begin{center}
		\includegraphics[width=1\linewidth]{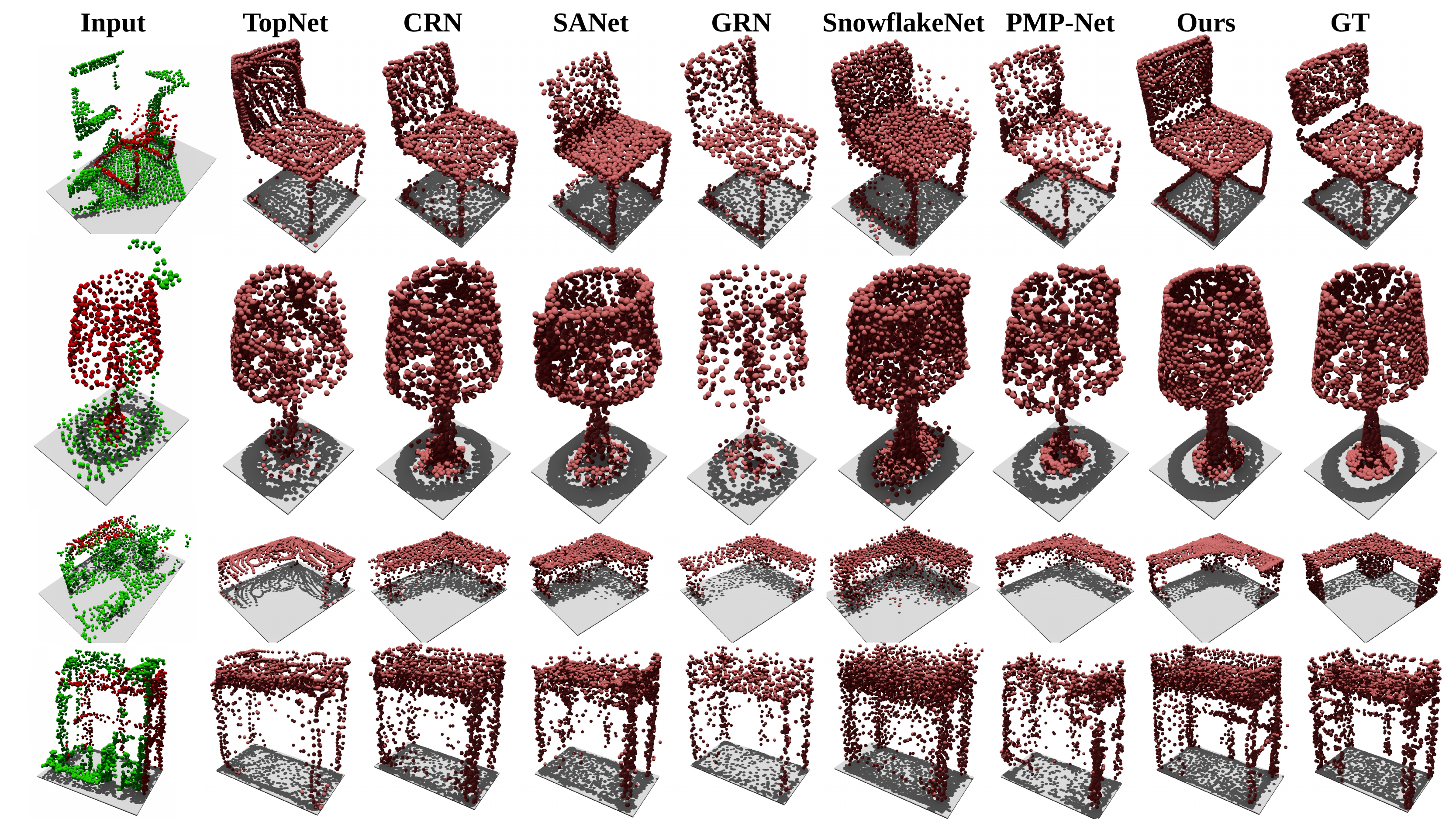}
	\end{center}
	\vspace{-2mm}
	\caption{Point cloud completion results by different methods. The categories from top to bottom: chair, lamp, table, and bed. Red and green in the inputs denote points on the partial objects and outliers, separately. The real inputs do not have the segmentation labels. }
	\label{res}
	\vspace{-2mm}
\end{figure*}

\section{Experiments}\label{sec:exp}


\subsection{Implementation details}
Our framework is implemented using PyTorch. The number of CS-Block $M$ is set to $3$ considering the performance, efficiency and training time. And the point number $N$ is set to $2048$. We train 10 categories altogether with the Adam optimizer. The learning rate of the optimizer and the batch size for our network are set to $1.2\times10^{-4}$ and 16 respectively. To ensure the convergence of neural networks, we train the model for around 120 epochs. Training our CS-Net takes about 20 hours for convergence with a GTX 2080Ti GPU. We will make our dataset and codes public in the future.

\begin{table}[t]
\renewcommand
\arraystretch{1}
\begin{center}
\caption{Comparison of completion results by our method and state-of-the-arts. The Chamfer Distance multiplied by \begin{math}10^{4}\end{math} is shown. The numbers of parameters of our method and state-of-the-arts are shown on the right.}
\vspace{-2mm}
\label{tab:sota}
\setlength{\tabcolsep}{2.5mm}{
\begin{tabular}{@{}r|cccc|c}
\toprule
Method      & CD $\downarrow$            & DCD $\downarrow$            & F$^{0.01\%}_{score}$ $\uparrow$ & F$^{0.1\%}_{score}$ $\uparrow$  & Para.(M) \\ \midrule
SnowflakeNet~\cite{xiang2021snowflakenet}& 10.79         & 0.558          & 0.248          & 0.908          & 19.32M     \\
PMP-Net~\cite{wen2021pmp}         & 9.34      & 0.668          & 0.245          & 0.914          & 5.44M     \\
TopNet~\cite{tchapmi2019topnet}      & 11.48         & 0.667          & 0.210          & 0.884          & 6.24M       \\
GRN~\cite{xie2020grnet}         & 10.68         & 0.785          & 0.177          & 0.885          & 76.72M     \\
SANet~\cite{wen2020point}       & 11.13         & 0.635          & 0.234          & 0.895          & 6.22M      \\
CRN~\cite{wang2020cascaded}         & 10.68         & 0.627          & 0.261          & 0.907          & 20.40M      \\ \midrule
\vspace{0.5mm}
Ours($\mathcal{P_C}_{pred}^{3}$) & \textbf{6.98} & \textbf{0.526} & \textbf{0.335} & \textbf{0.954} & 16.83M    \\
\vspace{0.5mm}
Ours($\mathcal{P_C}_{pred}^{2}$) & 7.56          & 0.554          & 0.306          & 0.947          &          \\
Ours($\mathcal{P_C}_{pred}^{1}$) & 12.13         & 0.667          & 0.201          & 0.874          &            \\ \bottomrule
\end{tabular}
}
\end{center}
\vspace{-8mm}
\end{table}

\subsection{Comparison}

For the first evaluation way mentioned in Section \ref{sec:dataset}, we compare our CS-Net against existing representative methods on point cloud completion, including TopNet~\cite{tchapmi2019topnet},  CRN~\cite{wang2020cascaded}, SA-Net~\cite{wen2020point}, GRN~\cite{xie2020grnet}, SnowflakeNet~\cite{xiang2021snowflakenet}, and PMP-Net~\cite{wen2021pmp}, both quantitatively and qualitatively. We retrain and test all the above methods using the above dataset we built for fair comparisons. Since these methods are trained with different object resolutions as outputs due to their network architectures, we sample the outputs of each method to 2048 points for fair comparisons. The comparison is shown in Table \ref{tab:sota}, where the average CD, DCD, F-Score@0.1\% and F-Score@0.01\% on 8 categories are listed. Our method shows superior performance on most object categories compared with previous methods. The average CD of ours is 25.2\% lower than PMP-Net \cite{wen2021pmp} which performs the best in all the methods used for comparison and represents state-of-the-arts. Table \ref{tab:sota} also shows the comparison of the completion results from three CS-Blocks.

\begin{figure}[t] 
	\begin{center}
		\includegraphics[width=0.5\linewidth]{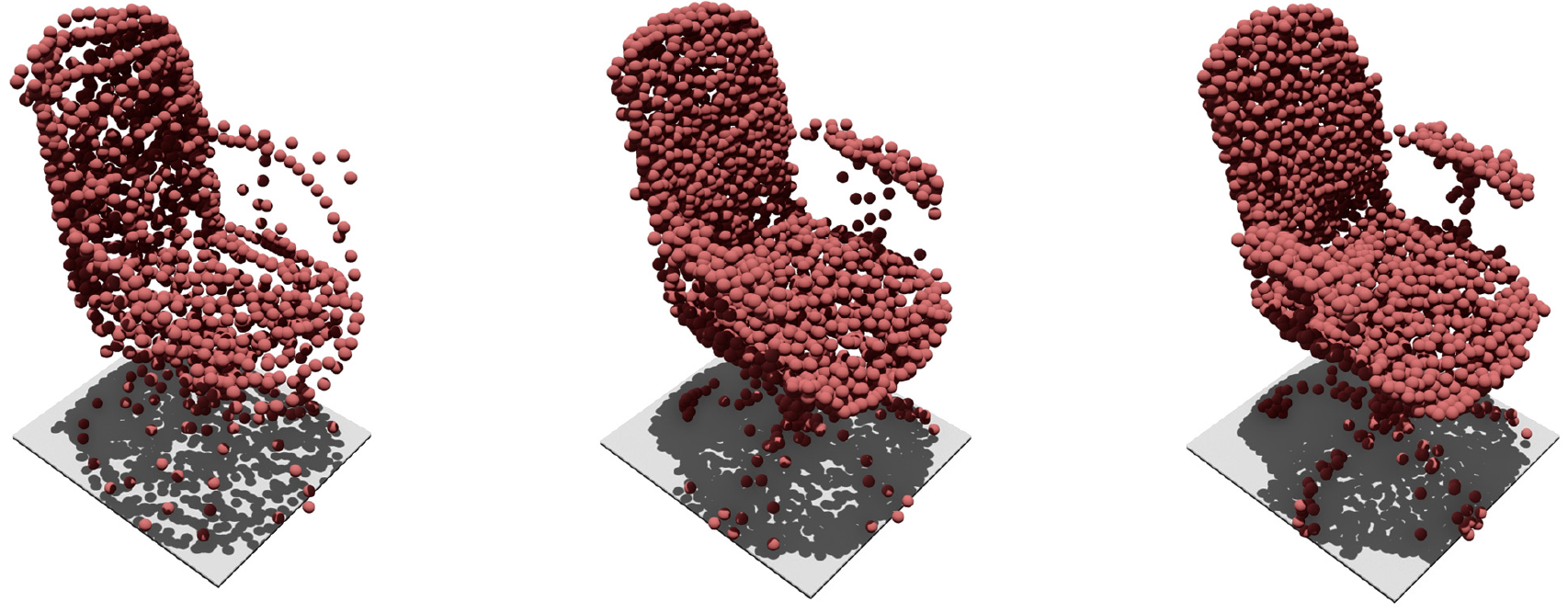}
	\end{center}
	\caption{Completion results of three CS-Blocks. From left to right are the results of the first block to the third block. A basic shape of chair can be seen in the result of the first block, but the point cloud still contains many noises. The second block removes more outliers and produces the point cloud with more uniform distribution, but with incorrect legs. The last block generates the finest details.}
	\label{net-steps}
\end{figure}

Qualitative results for the 4 categories are shown in Figure \ref{res}. For the chairs and tables in the first and third rows, the point clouds produced by our CS-Net present more uniform spatial distributions, which could be attributed to our label-multiplication-FPS module. Uniform point distribution would facilitate further applications including re-construction, simplification, and so on. The second and fourth rows show that CS-Net can predict complete point clouds with more accurate shapes and details. For the beds in the last row, the output of our CS-Net has fewer noises than others some of which even fill the bottom of the beds. Our network can also generate point clouds without being influenced by the outliers in the inputs. This can be shown in the second row where the outputs of others have bulges on the lampshades. These results also validate that our network is able to utilize the clean points from inputs and keep them in the final outputs such as the legs of the chairs, the lampshades, the corner of the desktops, and the crisscross boards of bed legs.
Figure \ref{net-steps} shows the results generated by CS-Blocks, which verifies that our cascaded architecture can refine the complete point cloud step by step. 

\begin{table}[t]
\begin{center}
\caption{Comparison of completion results by two-stage scheme framework and our method.}
\label{tab:rb_twsd}
\setlength{\tabcolsep}{2.5mm}{
\begin{tabular}{@{}c|c|cccc@{}}
\toprule
Seg.                 & Com.   & CD $\downarrow$           & DCD $\downarrow$           & F$^{0.01\%}_{score}$ $\uparrow$ & F$^{0.1\%}_{score}$ $\uparrow$      \\ \midrule
\multirow{6}{*}{PN2~\cite{2017PointNet++}} & Topnet~\cite{tchapmi2019topnet}  & 10.60         & 0.639          & 0.235          & 0.908          \\
                     & SANet~\cite{wen2020point}   & 12.14         & 0.650          & 0.223          & 0.895          \\
                     & GRN~\cite{xie2020grnet}     & 11.75         & 0.857          & 0.138          & 0.876          \\
                     & CRN~\cite{wang2020cascaded}     & 9.37          & 0.573          & 0.325          & 0.935          \\
                     & PMP-Net~\cite{wen2021pmp} & 9.82          & 0.623          & 0.297          & 0.923          \\
                     & SnowflakeNet~\cite{xiang2021snowflakenet}    & 10.68         & 0.527          & 0.281          & 0.925          \\ \midrule
\multirow{6}{*}{PT~\cite{zhao2021point}}  & Topnet~\cite{tchapmi2019topnet}  & 10.76         & 0.683          & 0.211          & 0.901          \\
                     & SANet~\cite{wen2020point}   & 10.49         & 0.633          & 0.237          & 0.907          \\
                     & GRN~\cite{xie2020grnet}     & 12.01         & 0.857          & 0.136          & 0.875          \\
                     & CRN~\cite{wang2020cascaded}      & 9.18          & 0.576          & 0.323          & 0.933          \\
                     & PMP-Net~\cite{wen2021pmp} & 9.70          & 0.631          & 0.285          & 0.914          \\
                     & SnowflakeNet~\cite{xiang2021snowflakenet}    & 9.31          & 0.556          & 0.316          & 0.935          \\ \midrule
                     \multicolumn{2}{c|}{Ours} & \textbf{6.98} & \textbf{0.526} & \textbf{0.335} & \textbf{0.954} \\ \bottomrule
\end{tabular}
}
\end{center}
\vspace{-8mm}
\end{table}

For the second evaluation way, we use PointNet++~\cite{2017PointNet++} and Point Transformer~\cite{zhao2021point} for segmentation and the above 6 networks for completion. As shown in Table \ref{tab:rb_twsd}, we evaluate the outcomes of the above two-stage scheme using the mean values of above metrics on 8 categories. And we use PN, PN2, PT denotes PointNet~\cite{2017PointNet},  PointNet++~\cite{2017PointNet++}, Point Transformer~\cite{zhao2021point} separately. Our CS-Net achieving simultaneous segmentation and completion in a unified network still shows superior performance compared with the two-stage scheme which runs segmentation and completion in two independent steps. For the two-stage scheme, since the segmentation network cannot always label the noises without any error, the inputs to completion sometimes contain outliers by which these completion methods are misdirected with poor fitting performance. This leads to the bad performance of some methods. 
By contrast, in our network, segmentation and completion work collaboratively, enabling mutual promotion with a cascaded structure. Our specially designed CS-Blocks also make the network refine the segmentation and completion results block by block.

\begin{table}[t]
\begin{center}
\caption{ Comparison with completion by state-of-the-art segmentation and completion networks trained with end-to-end framework. }
\label{tab:rb_ete}
\setlength{\tabcolsep}{2.5mm}{
\begin{tabular}{@{}c|c|cccc@{}}
\toprule
Com.                  & Seg. & CD $\downarrow$             & DCD $\downarrow$           & F$^{0.01\%}_{score}$ $\uparrow$ & F$^{0.1\%}_{score}$ $\uparrow$      \\ \midrule
\multirow{3}{*}{PMP-Net~\cite{wen2021pmp}}  & PN2~\cite{2017PointNet++}  & 10.12          & 0.677          & 0.232          & 0.899          \\
                      & PT~\cite{zhao2021point}  & 12.17          & 0.678          & 0.226          & 0.866          \\
                      & PSV~\cite{zhang2021pointsetvoting}   & 9.744          & 0.632          & 0.245          & 0.914          \\ \midrule
\multirow{3}{*}{SnowflakeNet~\cite{xiang2021snowflakenet}} & PN2~\cite{2017PointNet++}  & 11.72          & 0.574          & 0.228          & 0.897          \\
                      & PT~\cite{zhao2021point}  & 12.98          & 0.640          & 0.175          & 0.800          \\
                      & PSV~\cite{zhang2021pointsetvoting}   & 11.36          & 0.573          & 0.238          & 0.893          \\ \midrule
                      \multicolumn{2}{c|}{Ours} & \textbf{6.98} & \textbf{0.526} & \textbf{0.335} & \textbf{0.954} \\ \bottomrule
\end{tabular}
}
\end{center}
\vspace{-4mm}
\end{table}

\begin{table}[t]
\begin{center}
\caption{Comparison with state-of-the-art segmentation methods.}
\label{tab:rb_seg_res}
\setlength{\tabcolsep}{2.5mm}{
\begin{tabular}{@{}c|cccc@{}}
\toprule
Method & PSV~\cite{zhang2021pointsetvoting}   & PN2~\cite{2017PointNet++}  & PT~\cite{zhao2021point}    & Ours           \\ \midrule
mAcc. $\uparrow$    & 88.77 & 92.5 & 93.07 & \textbf{93.53} \\ \bottomrule
\end{tabular}
}
\end{center}
\vspace{-8mm}
\end{table}

For the third evaluation way, we use PSV~\cite{zhang2021pointsetvoting}, PointNet++~\cite{2017PointNet++} and Point Transformer~\cite{zhao2021point} as segmentation networks and SnowflakeNet~\cite{xiang2021snowflakenet} and PMP-Net~\cite{wen2021pmp} as completion networks for fusion, as shown in Table \ref{tab:rb_ete}. Because of the direct feature sharing between segmentation and completion in these end-to-end networks, the shared features need to offer the information for both segmentation and completion, which is hard for the network to learn, leading to the bad performance of the end-to-end framework.


We compare our CS-Net against existing representative methods on point cloud segmentation, including PointNet++~\cite{2017PointNet++}, PSV~\cite{zhang2021pointsetvoting} and Point Transformer~\cite{zhao2021point}. The mean accuracy of each method on 8 categories are shown in Table \ref{tab:rb_seg_res}. Note that these segmentation networks are usually designed for multi-classes segmentation, which may not suit for 2-classes segmentation task.


\subsection{Ablation Study}

\begin{table}[t]
\renewcommand
\arraystretch{1.2}
\begin{center}
\footnotesize
\caption{Ablation study of the proposed method.}
\label{tab:ablation}

\begin{tabular}{c|l|ccc}
\toprule
Index & Method & CD $\downarrow$& mAcc. $\uparrow$\\
\midrule
(1) & segmentation pipeline   & - & 93.13  \\
(2) & completion pipeline & 10.611 & - \\
(3) & (1) + (2)                          & 10.449 & 93.38  \\
(4) & (3) + label-multiplication-FPS                       & 7.427 & 93.49  \\
(5) & (4) + KNN-grouping refinement (full)             & \textbf{6.981} & 93.53  \\
(6) & 2-Block ($M=2$)         & 7.553 & 93.34 \\
(7) & 4-Block ($M=4$)          & 7.065 & \textbf{93.65} \\
\bottomrule
\end{tabular}
\end{center}
\vspace{-4mm}
\end{table}

\begin{table}[t]
\renewcommand
\arraystretch{1}
\begin{center}
\caption{Comparison of segmentation results obtained by applying different feature extractors independently and through integrating them into our framework.}
\label{tab:diff-en}

\setlength{\tabcolsep}{5mm}{
\begin{tabular}{l|cccc}
\toprule
method  & self & with ours \\
\midrule
PN~\cite{2017PointNet}        & 91.52             & 92.06  \\
PN2~\cite{2017PointNet++}         & \textbf{92.50}         & \textbf{93.53}  \\
PT~\cite{zhao2021point}  & 91.81   &  92.47  \\
\bottomrule
\end{tabular}}
\end{center}
\vspace{-8mm}
\end{table}

We conduct the ablation study to evaluate the necessity of important structures of our network including the combination of completion and segmentation, label-multiplication-FPS module, KNN-grouping refinement module and the number of CS-Block and analyze their effects. Table \ref{tab:ablation} compares different ablated networks trained on our dataset. The metrics are the average CD and segmentation accuracy. The networks of the first and second experiments are the independent segmentation branch and completion branch without sharing features. The third to fifth experiments gradually add the combination, the label-multiplication-FPS module, and the KNN-grouping refinement module to the independent segmentation and completion network. The network of the fifth one is the full CS-Net with $M=3$. The comparisons show the effect of these structures and prove the advancement of the collaboration between segmentation and completion. The last two experiments show the improvement of results with the number of CS-Block increasing. Since the gain brought by the four blocks is trivial, but at the expense of more memory and time consumption, the network with three blocks is the best considering both the accuracy and run-time performance. 

We also test the architectures that replace our feature extractor module in segmentation module with other backbones including PointNet~\cite{2017PointNet} , PointNet++~\cite{2017PointNet++}, and Point Transformer~\cite{zhao2021point}. As shown in Table \ref{tab:diff-en}, we compare the segmentation results of these extractors and with our CS-Net. The comparisons show that our architecture can accommodate different local feature extractors, improving their performance.


\section{Conclusions}
We have presented CS-Net, an end-to-end network specifically designed to complete point clouds contaminated by noises or containing outliers. Our network performs completion and segmentation simultaneously, which promote each other. This is realized with a cascaded network structure through which useful information extracted by the completion and segmentation branches is exchanged. The completion and segmentation results are gradually refined by several CS-Blocks. Our network is trained with a new dataset built upon public datasets which will be made publicly available to promote relevant research. We compare against previous methods to demonstrate our superiority. The comparison against the three evaluation methods also verifies our effectiveness. 

%
%
\bibliographystyle{splncs04}
\bibliography{egbib}
\end{document}